\documentclass[letterpaper, 10 pt, conference]{ieeeconf}  

\IEEEoverridecommandlockouts                           
\usepackage{microtype}
\usepackage{amsmath,amssymb,amsfonts}
\usepackage{algorithmic}
\usepackage{graphicx}
\usepackage{textcomp}
\usepackage{xcolor}
\usepackage{subfig}
\usepackage{gensymb}
\usepackage{caption}
\usepackage{hyperref}
\usepackage{mwe}
\usepackage{amsfonts}
\usepackage{amssymb}
\usepackage{verbatim}
\usepackage{amsbsy}
\usepackage{siunitx}
\usepackage{tabularx, booktabs}
\usepackage{dblfloatfix}
\usepackage{cite}
\usepackage{cleveref}

\usepackage{algorithm, algorithmic}
\usepackage[autolanguage]{numprint}

\usepackage{spreadtab}
\usepackage{multirow}
\usepackage{cite}
\usepackage{hyperref}
\usepackage{xcolor}
\usepackage{algorithmic}
\usepackage{textcomp}
\usepackage{xcolor}
\usepackage{subfig}
\usepackage{gensymb}
\usepackage{caption}
\usepackage{mwe}
\usepackage{amsfonts}
\usepackage{amssymb}
\usepackage{verbatim}
\usepackage{amsbsy}
\usepackage{siunitx}
\usepackage{tabularx, booktabs}
\usepackage{soul}
\usepackage{tikz}
\usepackage{bm}
\usetikzlibrary{calc}

\makeatletter
\newif\if@anonymize

\@anonymizetrue    

\if@anonymize
  \newcommand{\highlight@DoHighlight}{
    \fill [outer sep = -15pt, inner sep = 0pt, color=black]
          ($(begin highlight)+(0,8pt)$) rectangle ($(end highlight)+(0,-3pt)$) ;
  }

  \newcommand{\highlight@BeginHighlight}{
    \coordinate (begin highlight) at (0,0) ;
  }

  \newcommand{\highlight@EndHighlight}{
    \coordinate (end highlight) at (0,0) ;
  }

  \newdimen\highlight@previous
  \newdimen\highlight@current
  \newlength{\item@width}

  \DeclareRobustCommand*\anonymize{%
    \SOUL@setup
    \def\SOUL@preamble{%
      \begin{tikzpicture}[overlay, remember picture]
        \highlight@BeginHighlight
        \highlight@EndHighlight
      \end{tikzpicture}%
    }%
    \def\SOUL@postamble{%
      \begin{tikzpicture}[overlay, remember picture]
        \highlight@EndHighlight
        \highlight@DoHighlight
      \end{tikzpicture}%
    }%
    \def\SOUL@everyhyphen{%
      \discretionary{%
        \SOUL@setkern\SOUL@hyphkern
        \SOUL@sethyphenchar
        \tikz[overlay, remember picture] \highlight@EndHighlight ;%
      }{%
      }{%
        \SOUL@setkern\SOUL@charkern
      }%
    }%
    \def\SOUL@everyexhyphen##1{%
      \SOUL@setkern\SOUL@hyphkern
      \settowidth{\item@width}{##1}%
      \makebox[\item@width]{}%
      \discretionary{%
        \tikz[overlay, remember picture] \highlight@EndHighlight ;%
      }{%
      }{%
        \SOUL@setkern\SOUL@charkern
      }%
    }%
    \def\SOUL@everysyllable{%
      \begin{tikzpicture}[overlay, remember picture]
        \path let \p0 = (begin highlight), \p1 = (0,0) in \pgfextra
          \global\highlight@previous=\y0
          \global\highlight@current =\y1
        \endpgfextra (0,0) ;
        \ifdim\highlight@current < \highlight@previous
          \highlight@DoHighlight
          \highlight@BeginHighlight
        \fi
      \end{tikzpicture}%
      \settowidth{\item@width}{\the\SOUL@syllable}%
      \makebox[\item@width]{}%
      \tikz[overlay, remember picture] \highlight@EndHighlight ;%
    }%
    \SOUL@
  }
\else
  \newcommand{\anonymize}[1]{#1}
\fi
\makeatother 

\title{\LARGE \bf
Enhanced Low-Dimensional Sensing Mapless Navigation\\of Terrestrial Mobile Robots Using Double Deep\\Reinforcement Learning Techniques

}

\author{Linda Dotto de Moraes$^{1}$, Victor Augusto Kich$^{1}$, Alisson Henrique Kolling$^{1}$, Jair Augusto Bottega$^{1}$,\\\ Ricardo Bedin Grando$^{2}$, Anselmo Rafael Cukla$^{1}$, Daniel Fernando Tello Gamarra$^{1}$
\thanks{$^{1}$Linda Dotto de Moraes, Jair Augusto Bottega, Victor Augusto Kich, Alisson Henrique Kolling, Daniel Fernando Tello Gamarra and Anselmo Rafael Cukla are with Federal University of Santa Maria - UFSM, Santa Maria, RS, Brazil. E-mail: {\tt\small ricardo.bedin@utec.edu.uy}}
\thanks{$^{2}$Ricardo B. Grando is with the Universidad Tecnol\'ogica del Uruguay (UTEC). E-mail: {\tt\small ricardo.bedin@utec.edu.uy}}
}

\begin{document}

\maketitle
\let\thefootnote\relax\footnote{\\979-8-3503-1538-7/23/\$31.00\textcopyright2023
IEEE}

\thispagestyle{empty}
\pagestyle{empty}

\begin{abstract}

In this study, we present two distinct approaches of Deep Reinforcement Learning (Deep-RL) algorithms for a mobile robot. The research methodology primarily involves a comparative analysis between a Deep-RL strategy grounded in the foundational Deep Q-Network (DQN) algorithm, and the Double Deep Q-Network (DDQN) algorithm. The agents in these approaches leverage 24 measurements from laser range sampling, coupled with the agent's positional differentials and orientation relative to the target. This amalgamation of data influences the agents' determinations regarding navigation, ultimately dictating the robot's velocities. By embracing this parsimonious sensory framework as proposed, we successfully showcase the training of an agent for proficiently executing navigation tasks and adeptly circumventing obstacles. Notably, this accomplishment is attained without a dependency on intricate sensory inputs like those inherent to image-centric methodologies. The proposed methodology is evaluated in three different real environments, revealing that Double Deep structures significantly enhance the navigation capabilities of mobile robots compared to simple Q structures. 

\end{abstract}
\section*{SUPPLEMENTARY MATERIAL}

The experimental demonstrations are available at 
https://youtu.be/A29er5CGygw.
Released code, Docker image, and pre-trained models at
https://github.com/LindaMoraes/turtlebot-project.

\section{INTRODUCTION}



Reinforcement Learning (RL) provides a promising approach for a multitude of challenges in robotics.
 These approaches have showcased leading-edge effectiveness in addressing diverse challenges in robot learning scenarios, attributed to the progress made in the field of deep learning neural networks (Deep ANN). The portrayal of the agent through a Deep ANN has notably bolstered its competence in maneuvering intricate settings and accomplishing a spectrum of tasks. Nevertheless, this advancement has also ushered in novel complexities, particularly in the realm of learning from data sets with elevated dimensions. These intricacies arise due to the limitations set by ANN, including those linked to the learning gradient. To overcome this constraint, dedicated Deep-RL approaches such as Contrastive Learning have been utilized to mitigate the issue and streamline agent learning. Particularly noteworthy is the discovery that impressive results can be attained in navigation tasks by relying on basic sensory data. This fact has been validated in the context of ground-based mobile robots, aerial robots, underwater robots, and even hybrid robots. The principle of mapless navigation underpins various challenges in mobile robotics, leading to the successful application of numerous Deep-RL algorithms.

\begin{figure}[tbp!]
    \centering
    \includegraphics[width=\linewidth]{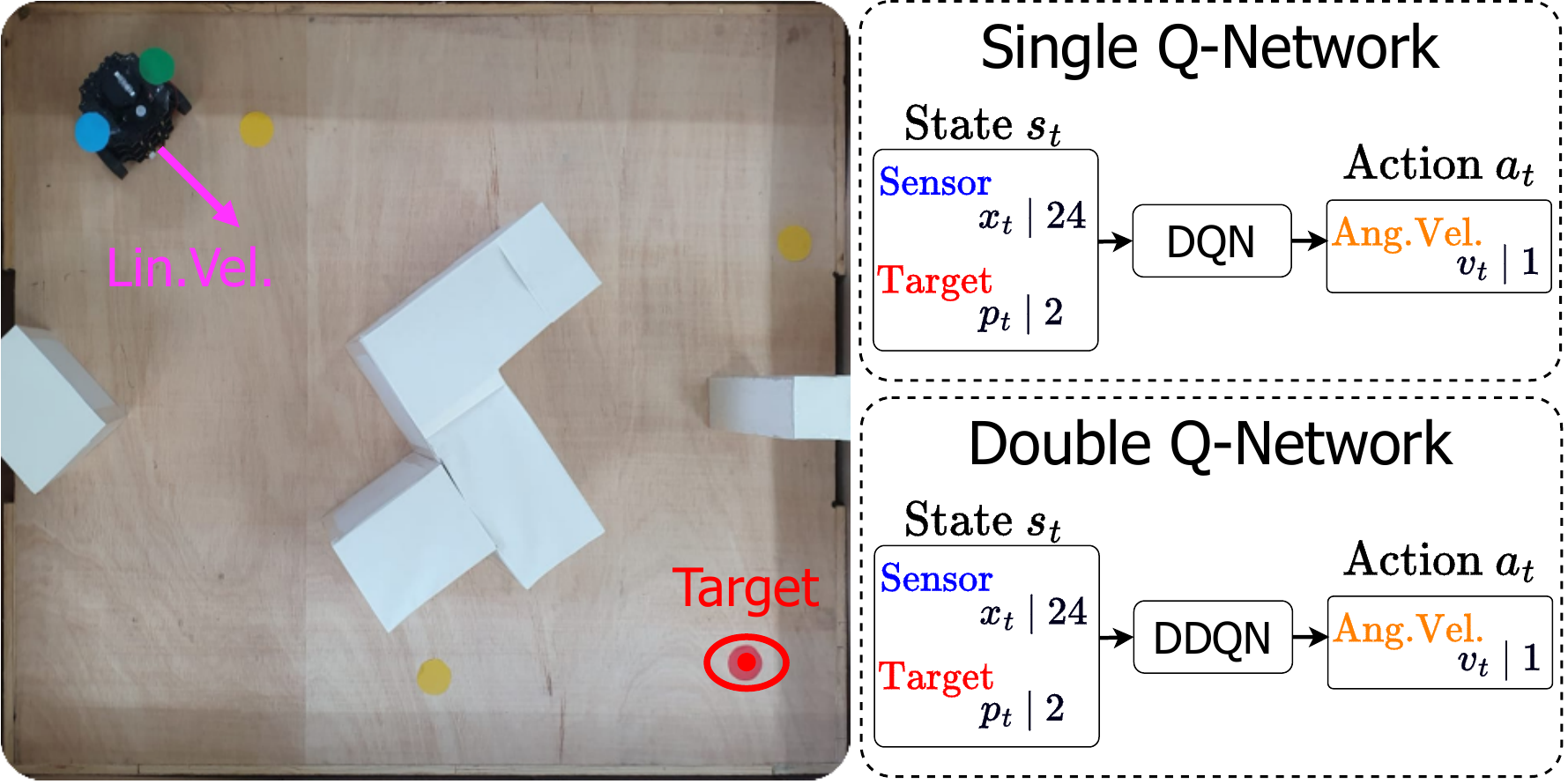}
    \caption{The Turtlebot3 Burger operates amidst obstacles in a specific scenario (on the left), while the model architecture, encompassing inputs and outputs of our techniques, namely DQN and Double DQN, is depicted on the right.}
    \label{fig:structure}
\end{figure}

Considering this perspective, this study aims to exhibit and assess the efficacy of two Deep-RL methodologies in tasks concerning the purpose-driven navigation of a ground-based mobile robot. A real-world evaluation of the DQN and DDQN algorithms is undertaken for comparative analysis. These strategies are rooted in the concept of straightforward sensing, wherein the design encompasses 26 state samples. This compilation comprises 24 readings from laser sensors, coupled with measurements of the mobile robot's distance and orientation relative to the target. Moreover, our emphasis extends to showcasing the performance distinction between Double Q architectures and conventional Q architectures. The architecture we propose for the learning process is illustrated in Figure \ref{fig:structure}.


Overall, this paper brings forth the subsequent contributions:

\begin{itemize}

\item The effectiveness of Double Q approaches in achieving mapless navigation for terrestrial mobile robots is demonstrated through real-world evaluations.

\item The integration of Double Q methodologies and a simplified sensing approach is demonstrated to be highly effective in tackling crucial obstacles within the domain of Deep-RL. This achievement encompasses pivotal challenges like the convergence intricacies of gradient descent and the persistent concern of catastrophic forgetting, providing a consistent and reliable means of mitigation.

\item A comprehensive framework is presented to facilitate future testing and exploration of Deep-RL approaches for mobile robots.

\end{itemize}

This manuscript comprises seven distinct sections. Commencing with a concise introduction, the subsequent segment (Section \ref{related_works}) delves into the studies conducted by fellow researchers in the domain, which have notably influenced and guided this present work. The subsequent portion (Section \ref{theoretical_background}) establishes the theoretical foundation for the algorithms that were implemented in the experimental phase. Subsequently, in Section \ref{expreimental_setup}, the suite of tools, software components, and environments harnessed for the study are elaborated upon. The instructional approach adopted to train the agent in achieving target objectives is elucidated in Section \ref{methodology}, accompanied by a comprehensive explication of the network architecture and the intricacies of the reward function employed. Ultimately, the outcomes and findings attained through this study are expounded upon in Section \ref{results}.
Lastly, the concluding section delves into the major achievements garnered and the potential applications of Deep-RL.

\section{RELATED WORK}\label{related_works}


Deep Reinforcement Learning has been applied previously to mapless navigation with mobile robots,
\cite{chen2017socially}, \cite{jesus2019deep}.
Mnih \textit{et al.} \cite{mnih2013playing} computed a value function for future reward using an algorithm called deep Q-network (DQN) for the Atari games, \cite{mnih2013playing}, \cite{van2016deep}.
It is important to emphasize that the DQN restricts itself to discrete actions when applied to a problem such as robot control.
In order to extend the DQN to continuous control, Lillicrap \textit{et al.} \cite{lillicrap2015continuous} proposes the deep deterministic policy gradients (DDPG) algorithm.
This innovation cleared the path for using Deep-RL in mobile robot navigation.


Tai \textit{et al.} \cite{tai2017virtual} pioneered a mapless motion planning solution for a mobile robot, utilizing sensor-derived range data and target position as system input to generate continuous steering commands. Their approach initially employed discrete steering commands\cite{tai2016towards}. Their study showcased the potential of training an agent, through asynchronous Deep-RL techniques, to achieve a predetermined target using this mapless motion planner.

Similarly inspired by Tai \textit{et al.} in \cite{tai2017virtual} and related works, the present paper centers on the creation of a mapless motion planning system based on low-dimensional range readings.
Diverging from previous approaches, our study uses a deterministic approach based on Double Deep Reinforcement Learning for solving navigation-related problems for a mobile robot and including a dynamic target for the terrestrial mobile robot in environments with no asynchronous training.
Overall, we demonstrate that low dimensional sensing data and simple Deep-RL approaches, such as the DDQN, can be used to excel at navigation-related tasks for terrestrial mobile robots.

Through this, we show that typical Deep-RL issues, such as the convergence of the gradient descent and the forgetting problems, can be effectively mitigated.

\section{THEORETICAL BACKGROUND}\label{theoretical_background}
\subsection{Deep Reinforcement Learning}
According to \cite{dai2019transformerxl},\cite{rao2019natural}, and \cite{tai2016towards}, with the advancements in deep learning, these techniques started being applied to methods that were previously inefficient. One of these methods was reinforcement learning, which could only be used for problems with limited sample sizes and relied on a linear function approximator. The capability of deep learning techniques to handle large volumes of data and inputs with high dimensionality aligns seamlessly with reinforcement learning methods, resulting in what is now known as deep reinforcement learning (Deep-RL) methods.

\subsection{Deep-Q Network - DQN}

Leading the recent breakthroughs in Deep-RL, we find the method known as Deep-Q Network (DQN) \cite{mnih2013playing}, which was developed by Mnih \textit{et al}. 
The DQN method incorporates key principles of reinforcement learning, including the utilization of the \textit{Bellman equation}, as described by:
\begin{equation}\label{eq:bellman}
    Q^{*}(s,a) = \mathbb{E}_{s'\sim\varepsilon}[r + \gamma \underset{a'}{max}Q^*(s',a')|s,a]
\end{equation}
An optimal state-action pair is given by the Bellman equation. By utilizing a neural network with weights $\theta$  that generates a function that calculates the action-value function, it is found $Q(s,a,\theta) \approx Q^*(s,a)$, ensuring convergence towards the optimal value. The training of this neural network implies minimizing the following equation:
\begin{equation}\label{eq:loss}
    \mathcal{L}(\theta) = [(y - Q(s,a,\theta))^2].
\end{equation}
With ${L}(\theta)$ named as loss function and $y  = \mathbb{E}_{s'\sim\varepsilon}[r + \gamma \underset{a'}{max}Q(s',a';\theta_{t})|s,a]$ being a target function derived from a network weights.

The neural network inputs are sampled state-action pairs derived from an experience replay buffer. The experience replay buffer stores each transition obtained by the agent. The agent has a $\epsilon$-greedy policy.

\subsection{Double Deep-Q Network - DDQN}

Hasselt \cite{hasselt2010double} proposed a solution known as Double Q-Learning. This algorithm calculates the next state value using a double estimator. 
Building upon the Double Q-Learning strategy in conjunction with DQN, Van Hasselt \textit{et al.} \cite{van2016deep}.
DDQN calculates the target function with the equation 
\begin{equation}
y_{t}^{DoubleDQN}  = r_{t+1} +  \gamma Q(S_{t+1}, \underset{a}{argmax}Q(s_{t+1},a;\theta_{t}),\theta_{t}^{-}),
\end{equation}


\section{Experimental Setup}\label{expreimental_setup}

This research involved conducting laboratory experiments using real robots, the main tools and experimental setup will be discussed.

\subsection{PyTorch}
The algorithms used in this work were written in Python
and the library PyTorch.
It is highly regarded for its user-friendly nature, simplicity, and integration of familiar Python concepts such as classes, structures, and conditional loops. The popularity of PyTorch has surged due to its performance and agility, aligning well with the demands of modern development. 
PyTorch's scalability is closely tied to its ease of use, efficiency, parallelism, and hardware acceleration. It has gained significant traction in commercial applications, with notable companies like Tesla, Facebook \cite{wu2019machine}, Uber, and many others adopting it. In academia, PyTorch is already extensively used in research fields such as natural language processing, image processing, object recognition, and more \cite{laskin2020curl}, \cite{dai2019transformerxl}, \cite{rao2019natural}.

\subsection{ROS}

ROS is considered a meta operating system that offers several standard services commonly associated with operating systems \cite{Emanueili2019}.
ROS adopts a graph architecture to represent the running processes, referred to as nodes, within the system. Communication between two or more processes is achieved through messages, which are exchanged over topics. 
In ROS, message exchange between nodes and topics follows the publishing and subscribing paradigm. Publishing involves sending data to a topic, while subscribing entails reading and receiving the data from that topic. ROS is particularly advantageous in applications that necessitate real-time sensor readings for decision-making by machines
\cite{deJesus2021}, \cite{ugalde2022}. To adhere to good development practices, it is recommended to create a new node for each new feature within the system.

\subsection{Gazebo}
Gazebo, an open-source 3D simulator \cite{fairchild2016ros}, serves as a valuable tool for conducting simulated experiments and greatly aids in the development process when used alongside ROS. It boasts a vast and thriving community encompassing academia, scientific research, and industry, which is currently experiencing rapid growth.
The utilization of Gazebo as a support tool is crucial during the initial stages of experimentation. It enables researchers to rapidly prototype and test ideas without the need for costly real-world implementations, which are often economically impractical in early development phases. The integration of Gazebo with ROS adds another layer of interest, as both tools are open source and benefit from highly active communities.
One of the most significant advantages of this integration lies in the ability to simulate various environments. Gazebo incorporates real-world rules and concepts, thereby facilitating the simulation of practical applications and scenarios.

\subsection{Turtlebot}



\begin{figure}[h]
    \centering
    \includegraphics[width=0.7\linewidth]{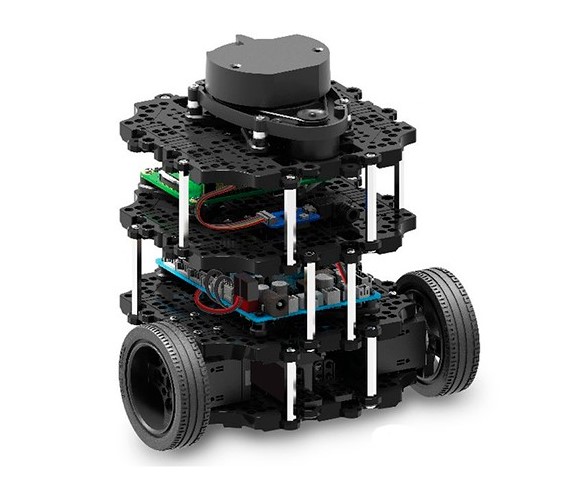}
    \caption{Turtlebot3 Burger.}
    \label{fig:turtlebot3}
\end{figure}

Within its product portfolio, Robotis offers the TurtleBot development kit, a line of educational robotics. This kit is renowned for its affordability in terms of hardware costs and utilization of open-source software, making it highly conducive for project implementation. 
The TurtleBot3 was used in the article has 
wheel encoders and a laser distance sensor. The adoption of Raspberry Pi3 further benefits from its widespread usage within the extensive community of TurtleBot users.  The turtlebot is shown in Figure \ref{fig:turtlebot3}.

\subsection{Experimental Environments}

After training the networks in simulation in Gazebo, real-world experiments were conducted using the TurtleBot in a physical environment. Overhead cameras captured images of the surroundings, which were subsequently processed using digital image processing algorithms. OpenCV, the most widely utilized open-source library for computer vision, was employed for image processing.

\begin{figure}[h]
  \subfloat[First stage.\label{fig:env_1_real}]{
	\begin{minipage}[c][\width]{0.15\textwidth}
	   \centering
	   \includegraphics[width=\textwidth]{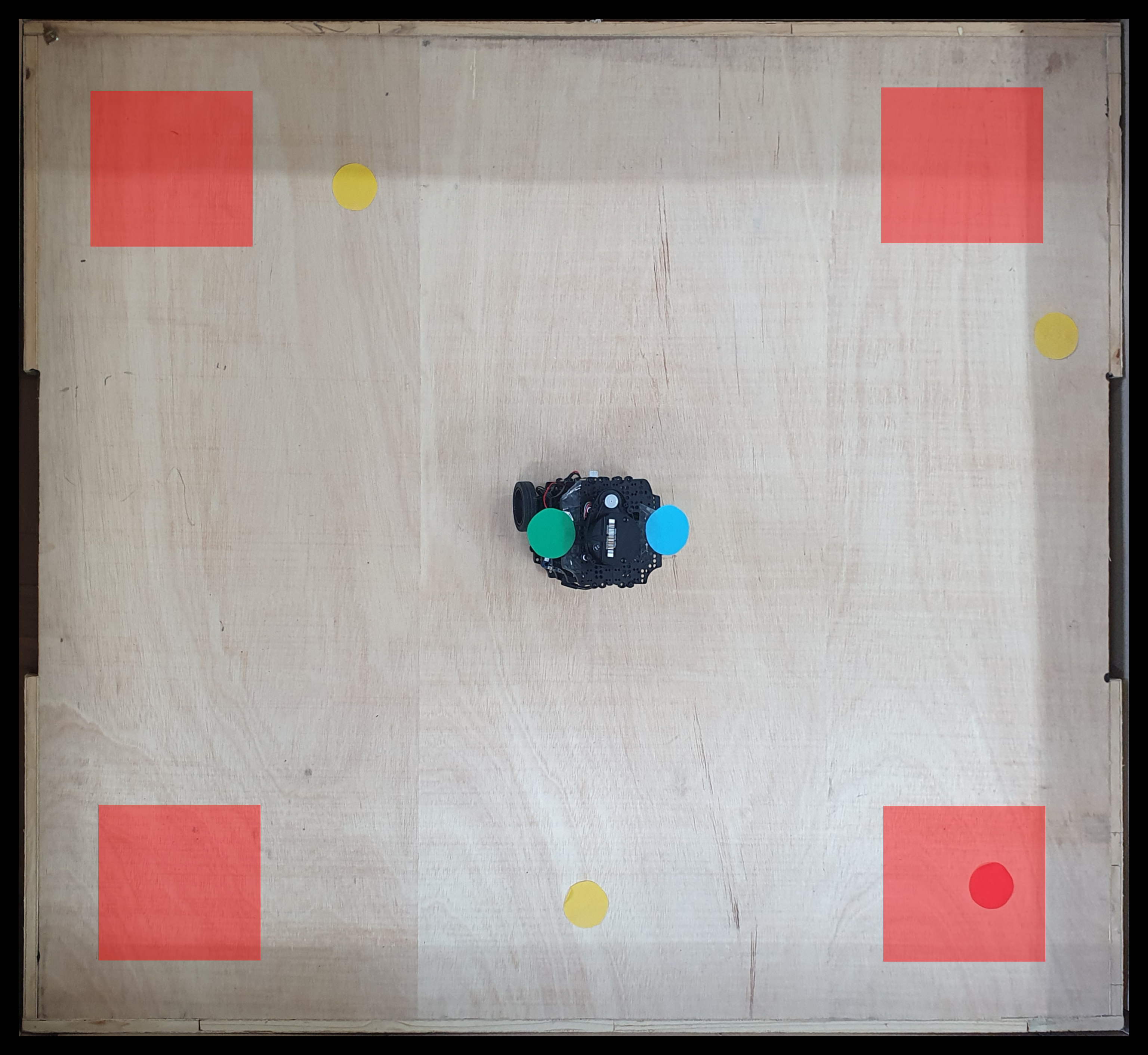}
	\end{minipage}}
 \hfill 	
  \subfloat[Second stage.\label{fig:env_2_real}]{
	\begin{minipage}[c][\width]{0.15\textwidth}
	   \centering
	   \includegraphics[width=\textwidth]{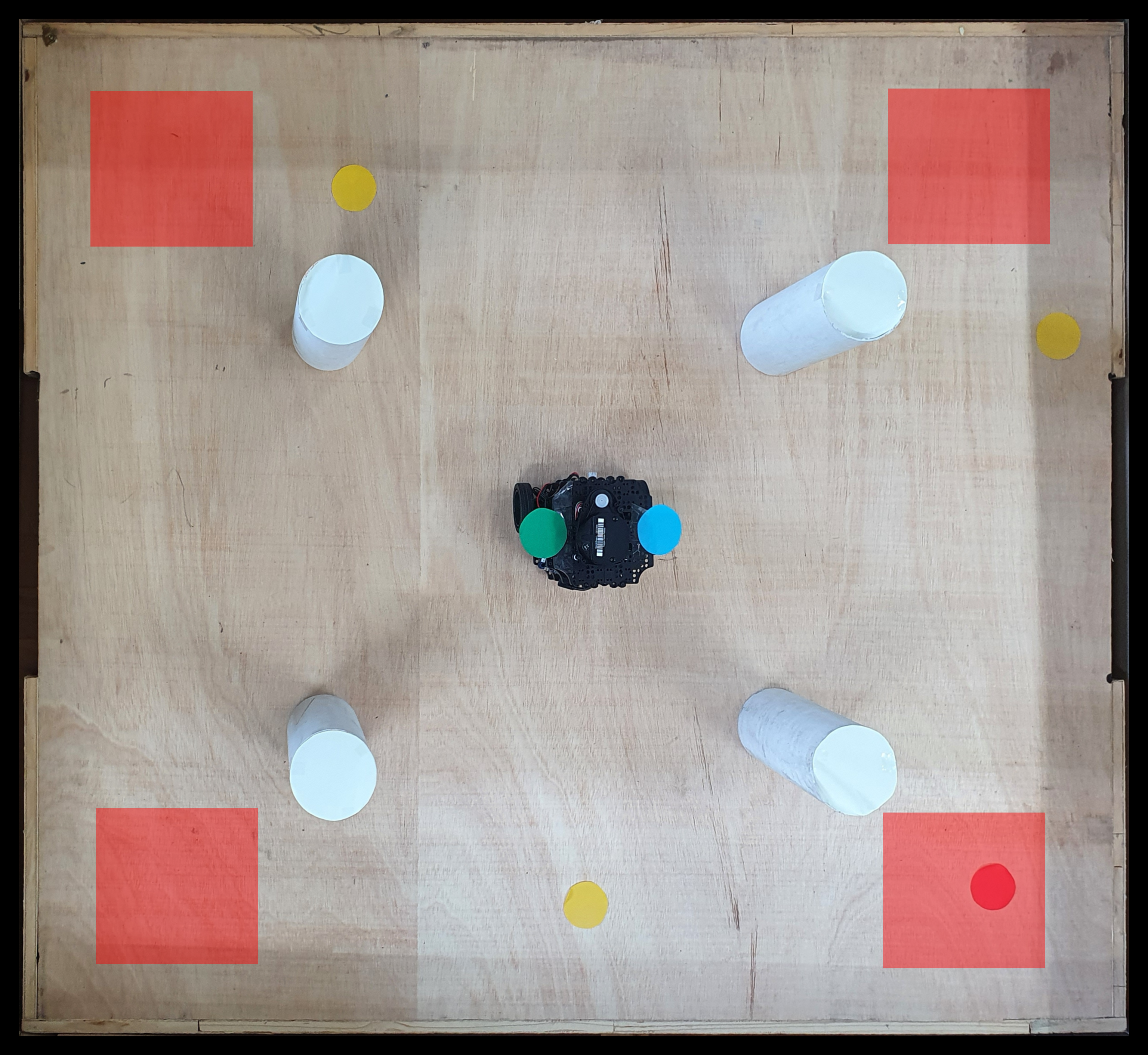}
	\end{minipage}}
 \hfill 
 \subfloat[Third stage.\label{fig:env_3_real}]{
	\begin{minipage}[c][\width]{0.15\textwidth}
	   \centering
	   \includegraphics[width=\textwidth]{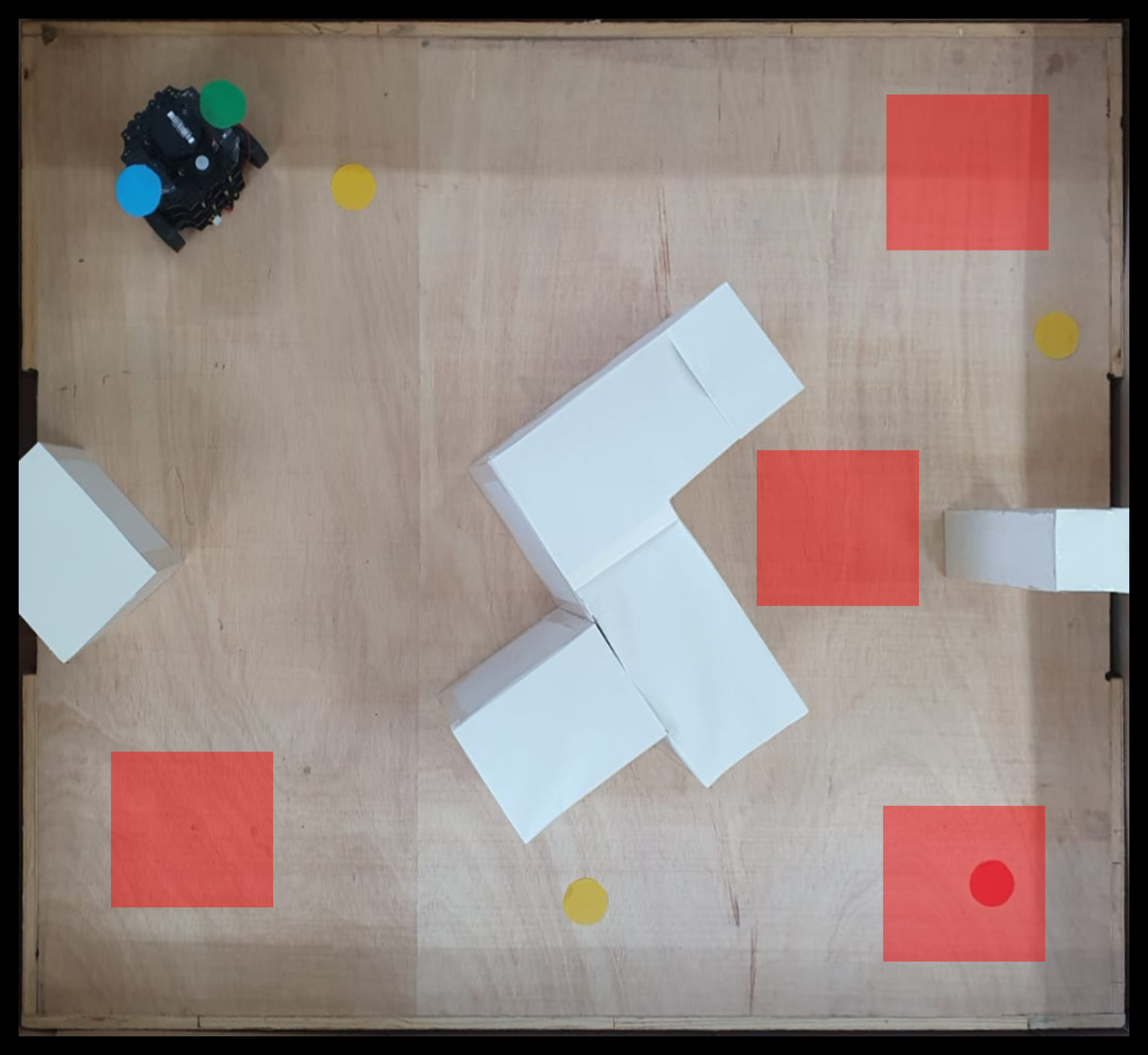}
	\end{minipage}}
\caption{Real experiment scenarios.}
\label{fig:environments_real}
\end{figure}

 The first real environment, as illustrated in Figure \ref{fig:env_1_real} does not have obstacles. Figure \ref{fig:env_2_real} showcases the second environment, which introduced a slightly higher level of difficulty compared to the initial scenario in which there are four obstacles. Lastly, Figure \ref{fig:env_3_real}, exhibits the third and final scenario, characterized by obstacles that have more complex geometry.

\section{METHODOLOGY}\label{methodology}



In this study, our objective is to train, test, and comprehensively compare the efficacy of the DQN and Double DQN algorithms when employed within the context of a Turtlebot3 platform.
The algorithms will let the robot navigate and avoid obstacles. The linear velocity is constant and the angular velocity has five discrete values.

\subsection{Network Structure}


\begin{figure}[t!]
    \centering
    \includegraphics[width=1\linewidth]{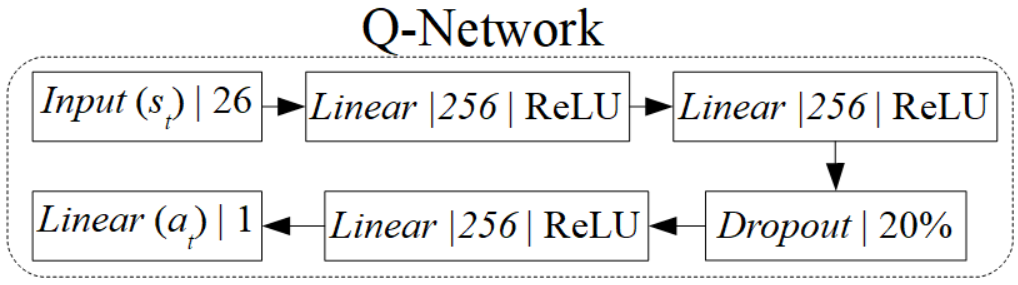}
    \caption{The Q-Network architecture applied in DQN and DDQN.}
    \label{fig:architecture}
\end{figure}

Once the system's states and actions have been defined, a Q-Network was developed to construct both the DQN and Double DQN architectures.
The network has 26 inputs that corresponds to readings of the laser sensor, previous angular and linear velocity, and position and orientation of the target.
The network's output corresponds to a discrete value within the range of [0,4], representing the angular velocity. Specifically, the values 0, 1, 2, 3, and 4 correspond to -1.5 rad/s, -0.75 rad/s, 0 rad/s, 0.75 rad/s, and 1.5 rad/s, respectively. 
Figure \ref{fig:architecture} shows the network architecture.

\begin{figure*}[t!]
  \subfloat[First stage.\label{fig:reward_1}]{
	\begin{minipage}[c][0.75\width]{0.32\textwidth}
	   \centering
	   \includegraphics[width=\textwidth]{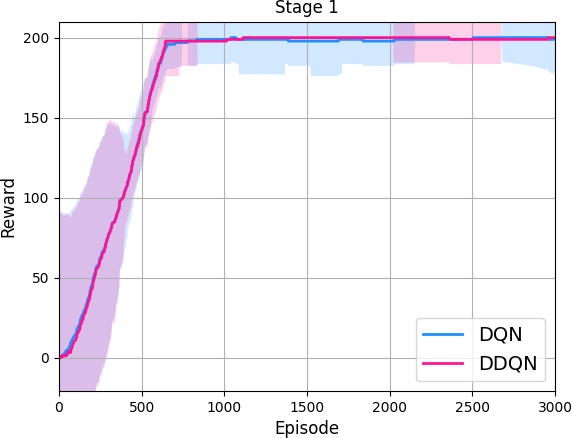}
	\end{minipage}}
 \hfill 	
  \subfloat[Second stage.\label{fig:reward_2}]{
	\begin{minipage}[c][0.75\width]{0.32\textwidth}
	   \centering
	   \includegraphics[width=\textwidth]{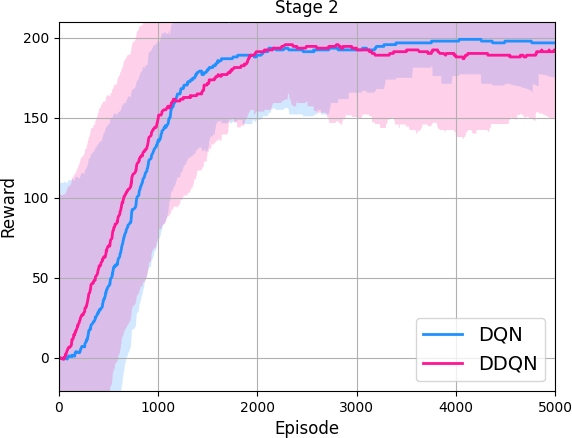}
	\end{minipage}}
 \hfill 
 \subfloat[Third stage.\label{fig:reward_3}]{
	\begin{minipage}[c][0.75\width]{0.32\textwidth}
	   \centering
	   \includegraphics[width=\textwidth]{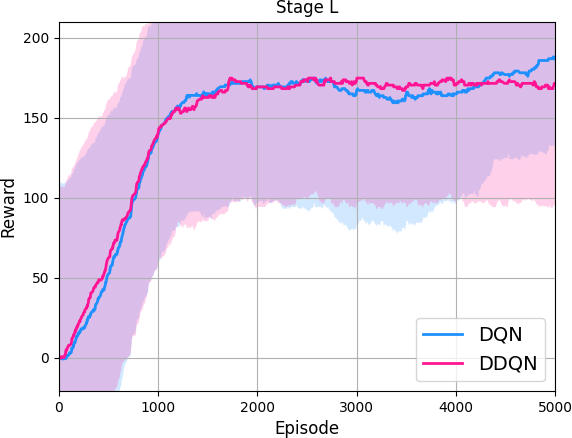}
	\end{minipage}}
\caption{Metrics of training episodes. DQN approach is represented by the blue line and DDQN by the pink.}
\label{fig:rewards}
\end{figure*}

The actor-network has three fully-connected layers with 256 nodes in each layer, the input of the network is the state of the robot. The output of the network is the angular velocities.
The linear velocity remains constant and predetermined at 0.15m/s, so there's no backward move on this configuration.

\begin{figure*}[bp]
  \subfloat[DQN in scenario 1.\label{fig:dqn_st1_real}]{
	\begin{minipage}[c][\width]{0.16\textwidth}
	   \centering
	   \includegraphics[width=\textwidth]{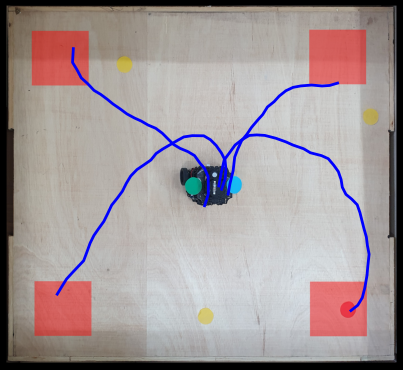}
	\end{minipage}}
  \subfloat[DDQN in scenario 1.\label{fig:ddqn_st1_real}]{
	\begin{minipage}[c][\width]{0.16\textwidth}
	   \centering
	   \includegraphics[width=\textwidth]{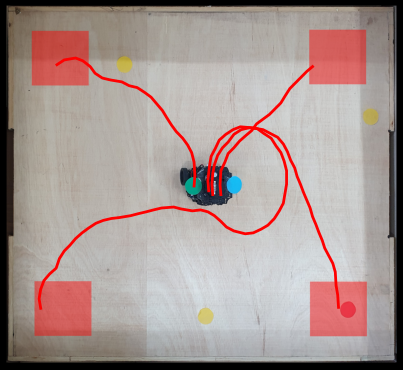}
	\end{minipage}}
 \subfloat[DQN in scenario 2.\label{fig:dqn_st2_real}]{
	\begin{minipage}[c][\width]{0.16\textwidth}
	   \centering
	   \includegraphics[width=\textwidth]{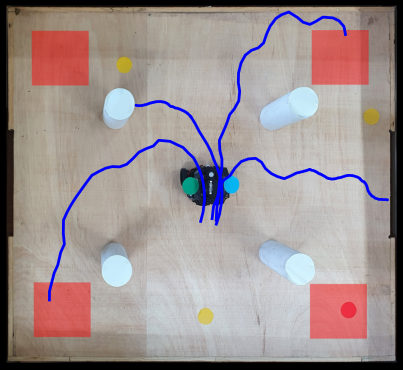}
	\end{minipage}}
  \subfloat[DDQN in scenario 2.\label{fig:ddqn_st2_real}]{
	\begin{minipage}[c][\width]{0.16\textwidth}
	   \centering
	   \includegraphics[width=\textwidth]{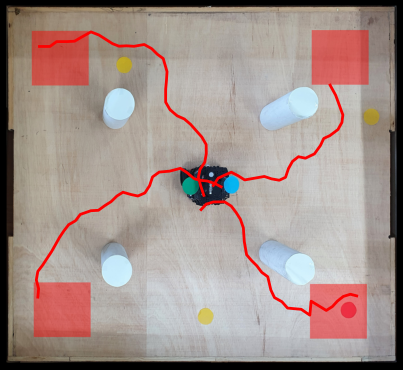}
	\end{minipage}}
  \subfloat[DQN in scenario 3.\label{fig:dqn_st3_real}]{
	\begin{minipage}[c][\width]{0.16\textwidth}
	   \centering
	   \includegraphics[width=\textwidth]{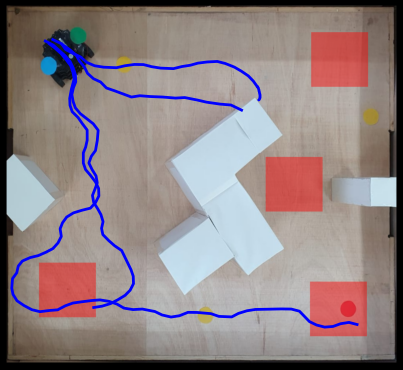}
	\end{minipage}}
  \subfloat[DDQN in scenario 3.\label{fig:ddqn_st3_real}]{
	\begin{minipage}[c][\width]{0.16\textwidth}
	   \centering
	   \includegraphics[width=\textwidth]{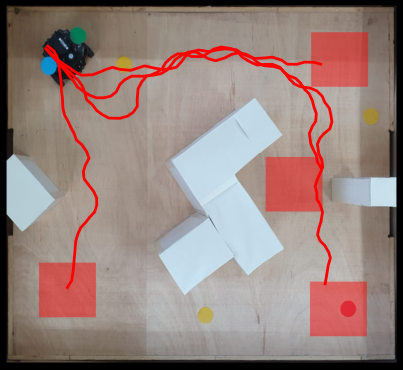}
	\end{minipage}}
\caption{The behavior of each approach was tested in each real scenario over 4 navigation trials.}
\label{fig:behavior_simulation_real}
\end{figure*}

\subsection{Reward Function}


The reward and penalty functions can be formulated based on empirical knowledge and developed iteratively during the problem-solving process. 

Regarding the reward system, the following three different conditions presented better results for the resolution of the problem:


\begin{equation}\label{reward_system}
r (s_t, a_t) = 
\begin{cases}
r_{arrive}\ \textrm{if} \ d_t < c_d\\
r_{collide}\ \textrm{if}\ min_x < c_o\\
r_{idle}\ \textrm{if}\ min_x >= c_o\ \textrm{and}\ d_t >= c_d
\end{cases}
\end{equation}



Only these three rewards were given, $r_{arrive}$ for making the task correctly, $r_{collide}$ in case of failure, and $r_{idle}$ in case of idle. A task is considered successful when the distance to goal ($d_t$) is less than the margin $c_d$, and the agent receives $200$ of reward denoted as $r_{arrive}$. This margin $c_d$, in this experiment, is set as $0.25$ meters. In the event of a collision against an obstacle or reaching the scenario boundaries, a negative reward $r_{collide}$ of $-20$ is given. A collision is determined by comparing the distance sensor readings to a threshold value $c_o$ of $0.12$. Additionally, if the agent maintains a distance $d_t<c_d$ from the target and its laser findings - expressed by $min_x$ - detect that the robot is keeping a distance upper or equal to $c_d$ from the obstacles and walls in a time period of 500 steps, the episode ends. In this last case,  a reward $r_{idle}$ of $0$ is given, and the episode is denominated idle since it didn't succeed nor collide.  
Simplifying the reward system into three conditions also helps focus on a more detailed examination of the Deep-RL approaches, their similarities, and differences rather than the intricacies of the scenario itself.


\section{RESULTS}\label{results}

\begin{table}[b]
\centering
\setlength{\tabcolsep}{3pt}
\caption{Episode Time, Sucess Rate, Standard and mean deviation metrics over $24$ navigation trials in three different real scenarios for DQN and DDQN approaches.}
\label{table:mean_std}
\begin{tabular}{c c c c} 
\toprule
\textbf{Env} & \textbf{Algorithm} & \bm{${ET}_{real}$} \textbf{(s)} & \bm{${SR}_{real}$} \\
\midrule
1 & DQN & \bm{$12.59$ $\pm$ $2.61$} & \bm{$100$}\textbf{\%} \\
1 & DDQN & $14.20$ $\pm$ $6.00$ & $100$\% \\
2 & DQN & $11.39$ $\pm$ $4.24$ & $50$\% \\
2 & DDQN & \bm{$20.11$ $\pm$ $5.88$} & \bm{$100$}\textbf{\%} \\
3 & DQN & $20.12$ $\pm$ $9.38$ & $50$\% \\
3 & DDQN & \bm{$22.28$ $\pm$ $6.69$} & \bm{$100$}\textbf{\%} \\
\bottomrule
\end{tabular}
\end{table}

This section presents the results obtained from this research. 
An extensive amount of statistical data was collected for each scenario and model.
The evaluation was done 
in a real workplace. 
A total of 24 test tasks were conducted, consisting of  4 iterations for each pre-trained model in each environment. Also, the trials were divided into four different fixed goals.
The number of successful trials was recorded, along with the average navigation time with their standard deviations. The Figure \ref{fig:rewards}, 
illustrates the learning in the training phase, showing metrics over 3000, 5000, and 5000 episodes respectively in each stage. Figure \ref{fig:behavior_simulation_real} provides the behavior of the robot during the evaluation. Furthermore, Table \ref{table:mean_std} presents the overall results gathered. 
Within Table \ref{table:mean_std}, we present the Episode Time (ET) and Success Rate (SR) corresponding to each test scenario.


An essential observation can be derived from Figure \ref{fig:rewards}, where the stable convergence of learning across the three evaluated scenarios is evident. 
However, the agent exhibited consistent stability throughout the learning process.

The robustness of these contributions is substantiated by the statistical analysis showcased in Table \ref{table:mean_std}, affirming the commendable performance exhibited by both algorithms.

The Double Q algorithm showcased an impressive performance, achieving nearly 100\% precision across diverse scenarios, thus underscoring its exceptional suitability for terrestrial mobile robotics.

Despite not being as sophisticated as contrastive learning algorithms or other recent Deep-RL approaches for continuous actions, the simplicity of the presented methodology exhibited the capability of reaching good performance levels.

\section{CONCLUSIONS}\label{conclusions}



This study introduces two straightforward Deep-RL techniques tailored to enhance the navigation capabilities of terrestrial mobile robots using low-dimensional data. Our results highlight the remarkable capabilities of the Double Q algorithms, showcasing their robust performance, which holds its ground even when juxtaposed with intricate Deep-RL methods such as actor-critic or contrastive architectures. The validation of these algorithms was carried out through real-world testing, underscoring their practical viability. Additionally, future testing of Deep-RL techniques can be scaled up using the framework that is provided in this work.

One noteworthy result unveiled by this paper is the coherent and steady learning trend that transcends various scenarios and temporal spans, with minimal signs of typical Deep-RL challenges like gradient convergence or memory loss. To provide additional weight to these conclusions, ongoing research endeavors are underway to validate this pattern across diverse categories of mobile robots and to delve into a spectrum of Deep-RL methodologies.

\section*{ACKNOWLEDGMENT}

We would like to thank the GARRA Research group $(https://www.ufsm.br/grupos/garra)$ and the VersusAI team. This work was partly founded by the Technological University of Uruguay (UTEC) and Federal University of Santa Maria (UFSM).

\bibliographystyle{./IEEEtran}
\bibliography{./main}

\end{document}